\documentclass[11pt,a4paper]{article}
\usepackage[utf8]{inputenc}

\usepackage[hyperref]{conll-2019}
\usepackage{times}
\usepackage{latexsym}

\usepackage{amsmath}

\usepackage{amssymb}

\usepackage{graphicx}

\usepackage{tabularx}
\usepackage{multirow}
\usepackage{booktabs}
\usepackage{diagbox}

\usepackage{bm}

\usepackage{todonotes}

\usepackage{comment}
\usepackage{multicol}

\makeatletter
\newcommand*\iftodonotes{\if@todonotes@disabled\expandafter\@secondoftwo\else\expandafter\@firstoftwo\fi}  
\makeatother

\newcommand{\modelname}{\textsc{butr}}

\usepackage{url}

\aclfinalcopy 

\title{Compositional Generalization in Image Captioning}
\renewcommand{\thefootnote}{\fnsymbol{footnote}}
\author{Mitja Nikolaus \\
  University of Tübingen\footnotemark[1] \\
  {\tt mitja.nikolaus@posteo.de} \\\And
  Mostafa Abdou \\
  University of Copenhagen \\
  {\tt abdou@di.ku.dk} \\\AND
  Matthew Lamm \\
  Stanford University \\
  {\tt mlamm@cs.stanford.edu} \\\And
  Rahul Aralikatte \\
  University of Copenhagen \\
  {\tt rahul@di.ku.dk} \\\And
  Desmond Elliott \\
  University of Copenhagen \\
  {\tt de@di.ku.dk} \\}

\date{}

\begin{document}
\maketitle

\begin{abstract}
Image captioning models are usually evaluated on their ability to describe a held-out set of images, not on their ability to generalize to unseen concepts. We study the problem of compositional generalization, which measures how well a model composes unseen combinations of concepts when describing images. State-of-the-art image captioning models show poor generalization performance on this task. We propose a multi-task model to address the poor performance, that combines caption generation and image--sentence ranking, and uses a decoding mechanism that re-ranks the captions according their similarity to the image. This model is substantially better at generalizing to unseen combinations of concepts compared to state-of-the-art captioning models.
\end{abstract}

\footnotetext[1]{The work was carried out during a visit to the University of Copenhagen.}
\renewcommand{\thefootnote}{\arabic{footnote}}

\section{Introduction}

When describing scenes, humans are able to almost arbitrarily combine concepts, producing novel combinations that they have not previously observed \cite{matthei1982acquisition, piantadosi2016compositional}. Imagine encountering a purple-colored dog in your town, for instance. Given that you understand the concepts \textsc{purple} and \textsc{dog}, you are able to compose them together to describe the dog in front of you, despite never having seen one before.

Image captioning models attempt to automatically describe scenes in natural language \cite{bernardi2016automatic}. Most recent approaches generate captions using a recurrent neural network, where the image is represented by features extracted from a Convolutional Neural Network (CNN). Although state-of-the-art models show good performance on challenge datasets, as measured by text-similarity metrics, their performance as measured by human judges is low when compared to human-written captions \cite[Section 5.3.2]{vinyals2017show}. 

\begin{figure}[!t]
\includegraphics[width=\columnwidth]{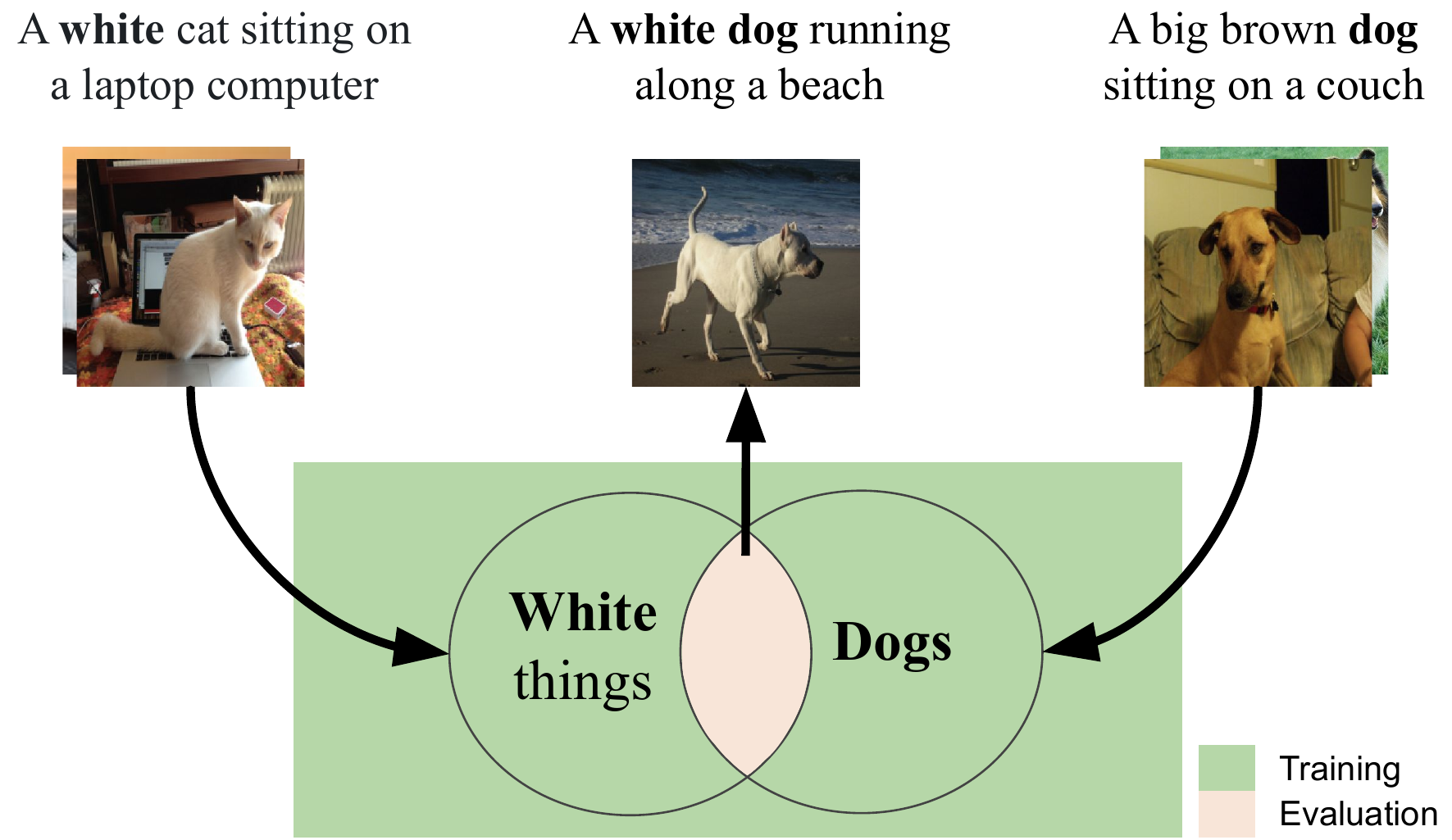}
\caption{\label{fig:intro} We evaluate whether image captioning models are able to compositionally generalize to unseen combinations of adjectives, nouns, and verbs by forcing \textit{paradigmatic gaps} in the training data.}
\end{figure}

 It is widely believed that systematic compositionality is a key property of human language that is essential for making generalizations from limited data \cite{montague1974formal, partee1984compositionality, lake2017building}. In this work, we investigate to what extent image captioning models are capable of compositional language understanding. We explore whether these models can compositionally generalize to unseen adjective--noun and noun--verb composition pairs, in which the constituents of the pair are observed during training but the combination is not, thus introducing a \textit{paradigmatic gap} in the training data, as illustrated in Figure \ref{fig:intro}. We define new training and evaluation splits of the COCO dataset \cite{chen2015microsoft} by holding out the data associated with the compositional pairs from the training set. These splits are used to evaluate how well models generalize to describing images that depict the held out pairings.

We find that state-of-the-art captioning models, such as Show, Attend and Tell \cite{Xu2015}, and Bottom-Up and Top-Down Attention \cite{anderson2018}, have poor compositional generalization performance.
We also observe that the inability to generalize of these models is primarily due to the language generation component, which relies too heavily on the distributional characteristics of the dataset and assigns low probabilities to unseen combinations of concepts in the evaluation data. This supports the findings from concurrent work \cite{holtzman2019curious} which studies the challenges in decoding from language models trained with a maximum likelihood objective. 

To address the generalization problem, we propose a multi-task model that jointly learns image captioning and image--sentence ranking. For caption generation, our model benefits from an additional step, where the set of captions generated by the model can be re-ranked using the jointly-trained image--sentence ranking component.
We find that the ranking component is less affected by the likelihood of $n$-gram sequences in the training data, and that it is able to assign a higher ranking to more informative captions which contain unseen combinations of concepts. These findings are reflected by improved compositional generalization. 

The source code is publicly available on GitHub.\footnote{\url{https://github.com/mitjanikolaus/compositional-image-captioning}}


\section{Related Work}

\subsection{Caption Generation and Retrieval}
\paragraph{Image Caption Generation} \hspace{-1em} models are usually end-to-end differentiable encoder-decoder models trained with a maximum likelihood objective. Given an image encoding that is extracted from a convolutional neural network (CNN), an RNN-based decoder generates a sequence of words that form the corresponding caption \cite[\textit{inter-alia}]{vinyals2015show}. This approach has been improved by applying top-down \cite{Xu2015} and bottom-up attention mechanisms \cite{anderson2018}.
These models show increasingly good performance on benchmark datasets, e.g. COCO, and in some cases reportedly surpass human-level performance as measured by n-gram based evaluation metrics \cite{bernardi2016automatic}. However, recent work has revealed several caveats. Firstly, when using human judgments for evaluation, the automatically generated captions are still considered worse in most cases \cite{fang2015captions,vinyals2017show}. Furthermore, when evaluating out-of-domain images or images with unseen concepts, it has been shown that the generated captions are often of poor quality \cite{mao2015learning, vinyals2017show}.
Attempts have been made to address the latter issue by leveraging unpaired text data or pre-trained language models \cite{anne2016deep, agrawal2018nocaps}.

\paragraph{Image--Sentence Ranking}\hspace{-1em} is closely related to image captioning. Here, the problem of language generation is circumvented and models are instead trained to rank a set of captions given an image, and vice-versa \cite{hodosh2013}. A common approach is to learn a visual--semantic embedding for the captions and images, and to rank the images or captions based on similarity in the joint embedding space. State-of-the-art models extract image features from CNNs and use gated RNNs to represent captions, both of which are projected into a joint space using a linear transformation \cite{frome2013devise, Karpathy2015Deep, Vendrov2016order, faghri2017}.

\subsection{Compositional Models of Language}
Investigations of compositionality in vector space models date back to early debates in the cognitive science \cite{fodor1988connectionism, fodor2002compositionality} and connectionist literature \cite{mcclelland1986parallel, smolensky1988proper} regarding the ability of connectionist systems to compose simple constituents into complex structures. In the NLP literature, numerous approaches that (loosely) follow the linguistic principle of compositionality\footnote{\textit{The meanings for complex expressions are derived from the meanings of their parts via specific composition functions.} \cite{partee1984compositionality}} have been proposed \cite{mitchell2008vector, baroni2010nouns, grefenstette2011experimental}. More recently, it has become standard to employ representations which are learned using neural network architectures. The extent to which these models behave compositionally is an open topic of research \cite{lake2017generalization, dasgupta2018evaluating, ettinger2018assessing, mccoy2018rnns} that closely relates to the focus of the present paper.

\paragraph{Compositional generalization in image captioning}\hspace{-1em} has received limited attention in the literature. In \citet{Atzmon2016}, the captions in the COCO dataset are replaced by subject-relation-object triplets, circumventing the problem of language generation, and replacing it with structured triplet prediction. Other work explores generalization to unseen combinations of visual concepts as a classification task \citep{misra2017red,kato2018compositional}.
\citet{lu2018neural} is more closely related to our work; they evaluate captioning models on describing images with unseen noun-noun pairs. 

In this paper, we study compositional generalization in image captioning with combinations of multiple classes of nouns, adjectives, and verbs.\footnote{This is different from the "robust image captioning" task \cite{lu2018neural} because we are testing for the composition of nouns with adjectives or verbs, and not the co-occurrence of different nouns in an image.} We find that state-of-the-art models fail to generalize to unseen combinations, and present a multi-task model that improves generalization by combining image captioning \cite{anderson2018} and image--sentence ranking \cite{faghri2017}. In contrast to other models that use a re-ranking step\footnote{\citet{fang2015captions} use a discriminative model that has access to sentence-level features and a multimodal similarity model in order to capture global semantics. \citet{wang2017diverse} uses a conditional variational auto-encoder to generate a set of diverse captions and a consensus-based method for re-ranking the candidates.}, our model is trained jointly on both tasks and does not use any additional features or external resources. The ranking model is only used to optimize the global semantics of the generated captions with respect to the image.

\section{Compositional Image Captioning}

\subsection{Problem Definition}
In this section we define the compositional captioning task, which is designed to evaluate how well a model generalizes to captioning images that \textit{should} be described using previously unseen combinations of concepts, when the individual concepts have been observed in the training data.

We assume a dataset of captioned images $\mathcal{D}$, in which $N$ images are described by $K$ captions: $\mathcal{D}$ $:=$ $\{\langle i^1,s^1_1,...,s^1_K\rangle, ..., \langle i^N,s^N_1,...,s^N_K\rangle\}$.
We also assume the existence of a concept pair $\{c_i, c_j\}$ that represents the concepts of interest in the evaluation.
In order to evaluate the compositional generalization of a model for that concept pair, we first define a training set by identifying and removing instances where the captions of an image contain the pair of concepts, creating a \textit{paradigmatic gap} in the original training set: $\mathcal{D}_\text{train} := \{\langle i^n,s^n_k\rangle\} \; \text{s.t.} \; \forall_{n=1}^{N} \nexists \; k: c_i \in s^n_k \land c_j \in s^n_k$. Note that the concepts $c_i$ and $c_j$ can still be independently observed in the captions of an image of this set, but \textit{not} together in the same caption.
We also define validation and evaluation sets $D_{val}$ and $D_{eval}$ that \textit{only} contain instances where at least one of the captions of an image contains the pair of concepts:
$\mathcal{D}_\text{val/eval} := \{\langle i^n,s^n_k\rangle\} \; \text{s.t.} \; \forall_{n=1}^{N} \exists \; k: c_i \in s^n_k \land c_j \in s^n_k$. 
A model is trained on the $\mathcal{D}_\text{train}$ training set until it converges, as measured on the $\mathcal{D}_\text{val}$ validation set. The compositional generalization of the model is measured by the proportion of evaluation set captions which successfully combined a held out pair of concepts $\{c_i, c_j\}$ in $D_{eval}$.

\subsection{Selection of Concept Pairs}
We select pairs of concepts that are likely to be represented in an image recognition model. In particular, we identify adjectives, nouns, and verbs in the English COCO captions dataset \cite{chen2015microsoft} that are suitable for testing compositional generalization. 
We define concepts as sets of synonyms for each word, to account for the variation in how the concept can be expressed in a caption. For each noun, we use the synonyms defined in \citet{lu2018neural}. For the verbs and adjectives, we use manually defined synonyms (see Appendix \ref{appendix:synonyms}). From these concepts, we select adjective--noun and noun--verb pairs for the evaluation. To identify concept pair candidates, we use StanfordNLP \cite{qi2018} to label and lemmatize the nouns, adjectives, and verbs in the captions, and to check if the adjective or verb is connected to the respective noun in the dependency parse.

\paragraph{Nouns:} We consider the 80 COCO object categories \cite{lin2014} and additionally divide the ``person" category into ``man", ``woman" and ``child". It has been shown that models can detect and classify these categories with high confidence \cite{he2016deep}. We further group the nouns under consideration into animate and inanimate objects. We use the following nouns in the evaluation: \texttt{woman}, \texttt{man},  \texttt{dog}, \texttt{cat}, \texttt{horse}, \texttt{bird}, \texttt{child}, \texttt{bus}, \texttt{plane}, \texttt{truck}, \texttt{table}.

\paragraph{Adjectives:} We analyze the distribution of the adjectives in the dataset (see Figure \ref{fig:adjectives} in Appendix A). The captions most frequently contain descriptions of the color, size, age, texture or quantity of objects in the images. We consider the color and size adjectives in this evaluation. It has been shown that CNNs can accurately classify the color of objects \cite{anderson2016}; and we assume that CNNs can encode the size of objects because they can predict bounding boxes, even for small objects \cite{bai2018sod}. In the evaluation, we use the following adjectives: \texttt{big}, \texttt{small}, \texttt{black}, \texttt{red}, \texttt{brown}, \texttt{white}, \texttt{blue}.

\paragraph{Verbs:} \citet{sadeghi2011recognition} show that it is possible to automatically describe the interaction of objects or the activities of objects in images. We select verbs that describe simple and well-defined actions and group them into transitive and intransitive verbs.
 We use the following verbs in the pairs: \texttt{eat}, \texttt{lie}, \texttt{ride}, \texttt{fly}, \texttt{hold}, \texttt{stand}.

\begin{table}
    \centering
    \small{
    \begin{tabular}{lll}
         \cmidrule(lr){1-3}
         \texttt{black cat} & \texttt{big bird} & \texttt{red bus}  \\
         \texttt{small plane} & \texttt{eat man} & \texttt{lie woman} \\
         \texttt{white truck} & \texttt{small cat} & \texttt{brown dog}  \\
         \texttt{big plane} & \texttt{ride woman} & \texttt{fly bird} \\
         \texttt{white horse} & \texttt{big cat} & \texttt{blue bus} \\
         \texttt{small table} & \texttt{hold child} & \texttt{stand bird} \\
         \texttt{black bird} & \texttt{small dog} & \texttt{white boat} \\
         \texttt{stand child} & \texttt{big truck} & \texttt{eat horse}\\
         \cmidrule(lr){1-3}
    \end{tabular}
    }
    \caption{The 24 concept pairs used to construct the training $\mathcal{D}_\textrm{train}$ and eval $\mathcal{D}_\textrm{eval}$ datasets.}
    \label{tab:task:word_pairs}
\end{table}

\paragraph{Pairs and Datasets:} We define a total of 24 concept pairs for the evaluation, as shown in Table \ref{tab:task:word_pairs}. The training and evaluation data is extracted from the COCO dataset, which contains $K$=5 reference captions for $N$=123,287 images. In the compositional captioning evaluation, we define the training datasets $D_{train}$ and validation datasets $D_{val}$ as subsets of the original COCO training data, and the evaluation datasets $D_{eval}$ as subsets of the COCO validation set, both given the concept pairs.

To ensure that there is enough evaluation data, we only use concept pairs for which there are more than 100 instances in the validation set. Occurrence statistics for the considered concept pairs can be found in Appendix \ref{appendix:held_out_pairs}.

\subsection{Evaluation Metric}\label{sec:evaluation_metric}
The performance of a model is measured on the $\mathcal{D}_\textrm{eval}$ datasets. For each concept pair evaluation set consisting of $M$ images, we dependency parse the set of $M$ $\times$ $K$ generated captions $\{\langle s^1_1,...,s^1_K\rangle, ..., \langle s^M_1,...,s^M_K\rangle\}$ to determine whether the captions contain the expected concept pair, and whether the adjective or verb is a dependent of the noun.\footnote{This means that a model gains no credit for predicting the concept pairs without them attaching to their expected target.} We denote the set of captions for which these conditions hold true as $\mathcal{C}$.

There is low inter-annotator agreement in the human reference captions on the usage of the concepts in the target pairs.\footnote{\label{footnote:inter_annotator_agreement}We calculate the inter-annotator agreement for the target pairs between the 5 reference captions for every image in the COCO dataset: on average, only 1.57 / 5 captions contain the respective adjective--noun or noun--verb concept pair, if it is present in any. We ascribe this lack of agreement to the open nature of the annotation task: there were no restrictions given for what should be included in an image caption.} Therefore, one should \textit{not} expect a model to generate a \textit{single} caption with the concepts in a pair. However, a model can generate a larger set of $K$ captions using beam search or diverse decoding strategies. Given $K$ captions, the recall of the concept pairs in an evaluation dataset is:
\begin{equation}\label{eq:metric}
    \textrm{Recall@K} = \frac{|\{\langle s^m_k \rangle \; | \; \exists k: s^m_k \in \mathcal{C} \}|}{M}
\end{equation}

Recall@K is an appropriate metric because the reference captions were produced by annotators who did not need to produce any specific word when describing an image. In addition, the set of captions $\mathcal{C}$ is determined with respect to the same synonym sets of the concepts that were used to construct the datasets, and so credit is given for semantically equivalent outputs. More exhaustive approaches to determine semantic equivalence for this metric are left for future work.

\section{State-of-the-Art Performance}\label{sec:experiments}

\subsection{Experimental Protocol}

\paragraph{Models:} \hspace{-1em} We evaluate two image captioning models on the compositional generalization task: Show, Attend and Tell \cite[\textsc{sat};][]{Xu2015} and Bottom-up and Top-down Attention \cite[\textsc{butd};][]{anderson2018}. For \textsc{sat}, we use ResNet-152 \cite{he2016deep} as an improved image encoder.

\paragraph{Training and Evaluation:}
The models are trained on the $\mathcal{D}_{\textrm{train}}$ datasets, in which groups of concept pairs are held out---see Appendix \ref{appendix:dataset_splits} for more information. Hyperparameters are set as described in the respective papers.
When a model has converged on the $\mathcal{D}_{\textrm{val}}$ validation split (as measured in BLEU score), we generate $K$ captions for each image in $\mathcal{D}_{\textrm{eval}}$ using beam search. Then, we calculate the Recall@K metric (Eqn. \ref{eq:metric}, K=5) for each concept pair in the evaluation split, as well as the average over all recall scores to report the compositional generalization performance of a model.

We also evaluate the compositional generalization of a \textsc{butd} model trained on the full COCO training dataset (\textsc{full}). In this setting, the model is trained on compositions of the type we seek to evaluate in this task, and thus does not need to generalize to new compositions.

\paragraph{Pretrained Language Representations:}
The word embeddings of image captioning models are usually learned from scratch, without pre-training\footnote{Exceptions: \citet{you2016image, anderson2017guided}}.
Pretrained word embeddings (e.g. GloVe \citep{pennington2014glove}) or language models (e.g. \citet{devlin2018bert}) contain distributional information obtained from large-scale textual resources, which may improve generalization performance. However, we do use them for this task because the resulting model may not have the expected paradigmatic gaps.

\subsection{Results}

\paragraph{Image Captioning:} The models mostly fail to generate captions that contain the held out pairs. The average Recall@5 for \textsc{sat} and \textsc{butd} are 3.0 and 6.5, respectively.
A qualitative analysis of the generated captions shows that the models usually describe the depicted objects correctly, but, in the case of held out adjective--noun pairs, the models either avoid using adjectives, or use adjectives that describe a different property of the object in question, e.g. \textit{white and green airplane} instead of \textit{small plane} in Figure \ref{fig:example_captions}.
In the case of held out noun--verb pairs, the models either replace the target verb with a less descriptive phrase, e.g. \textit{a man sitting with a plate of food} instead of \textit{a man is eating} in Figure \ref{fig:example_captions}, or completely omit the verb, reducing the caption to a simple noun phrase.

In the \textsc{full} setting, average Recall@5 reaches 33.3. We assume that this score is a conservative estimate due to the low average inter-annotator agreement (see Footnote \ref{footnote:inter_annotator_agreement}). The model is less likely to describe an image using the target pair if the pair is only present in one of the reference captions, as the feature is likely not salient (e.g. the car in the image has multiple colors, and the target color is only covering one part of the car). In fact, if we calculate the average recall for images where at least 2 / 3 / 4 / 5 of the reference captions contain the target concept pair, Recall@5 increases to 46.5 / 58.3 / 64.9 / 75.2. This shows that the \textsc{butd} model is more likely to generate a caption with the expected concept pair when more human annotators agree that it is a salient pair of concepts in an image.

\paragraph{Image--Sentence Ranking:}
In a related experiment, we evaluate the generalization performance of the VSE++ image--sentence ranking model on the compositional captioning task \cite{faghri2017}. We use an adapted version of the evaluation metric because the ranking model does not generate tokens.\footnote{For each image in the evaluation set, we construct a test set that consists of the 5 correct captions and the captions of 1,000 randomly selected images from the COCO validation set. We ensure that all captions in the test set contain exactly one of the constituent concept pairs, but not both (except for the 5 correct captions). We construct a ranking of the captions in this test set with respect to the image, and use the top-$K$ ranked captions to calculate the concept pair recall (Eqn. \ref{eq:metric}).}
The average Recall@5 with the adapted metric for the ranking model is 46.3. The respective \textsc{full} performance for this model is 47.0, indicating that the model performs well whether it has seen examples of the evaluation concept pair at training time or not. In other words, the model achieves better compositional generalization than the captioning models.

\section{Joint Model}\label{sec:model}

In the previous section, we found that state-of-the-art captioning models fail to generalize to unseen combinations of concepts, however, an image-sentence ranking model does generalize. We propose a multi-task model that is trained for image captioning and image--sentence ranking with shared parameters between the different tasks. The captioning component can use the ranking component to re-rank complete candidate captions in the beam. This ensures that the generated captions are as informative and accurate as possible, given the constraints of satisfying both tasks.

\begin{figure*}[!htb]
\center{\includegraphics[width=0.7\textwidth] 
{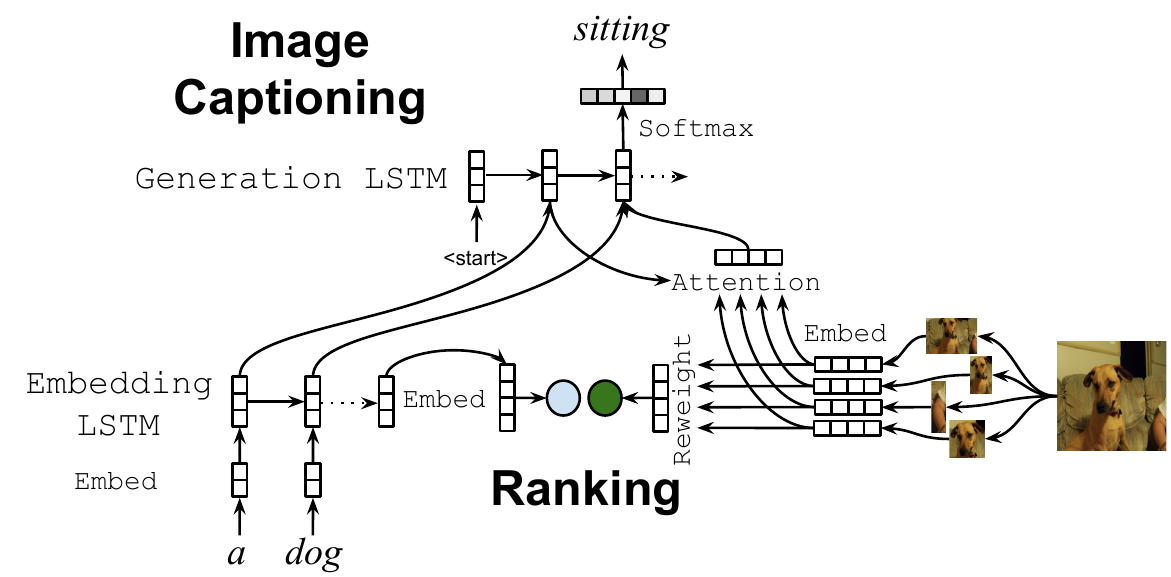}}
\caption{\label{fig:high_level_arch} An overview of \modelname{}, which jointly learns image--sentence ranking and image captioning.}
\end{figure*}

Following \citet{anderson2018}, the model is a two-layer LSTM \cite{hochreiter1997long}, where the first layer encodes the sequence of words, and the second layer integrates visual features from the bottom-up and top-down attention mechanism, and generates the output sequence. The parameters of the ranking component $\theta_{2}$ are mostly a subset of the parameters of the generation component $\theta_{1}$. We name the model \textbf{B}ottom-\textbf{U}p and \textbf{T}op-down attention with \textbf{R}anking (\modelname). 
Figure \ref{fig:high_level_arch} shows a high-level overview of the model architecture. 

\subsection{Image--Sentence Ranking}
To perform the image--sentence ranking task, we project the images and captions into a joint visual-semantic embedding space $\mathbb{R}^{J}$.
We introduce a language encoding LSTM with a hidden layer dimension of $L$.
\begin{equation} \label{eq:lang_enc_lstm}
    \bm{h}^l_t = \text{LSTM}(\bm{W}_1 \bm{o}_t, \bm{h}^l_{t-1})
\end{equation}
where $\bm{o}_t \in \mathbb{R}^{V}$ is a one-hot encoding of the input word at timestep $t$, $\bm{W}_1 \in \mathbb{R}^{E \times V}$ is a word embedding matrix for a vocabulary of size $V$ and $\bm{h}^l_{t-1}$ the state of the LSTM at the previous timestep. At training time, the input words are the words of the target caption at each timestep.

The final hidden state of the language encoding LSTM $\bm{h}^l_{t=T}$ is projected into the joint embedding space as $\bm{s}^* \in \mathbb{R}^{J}$ using $\bm{W}_2 \in \mathbb{R}^{J \times L}$:
\begin{align}
\label{eq:projected_encoder_embeddings}
    \bm{s}^* & = \bm{W}_2 \bm{h}^l_{t=T}
\end{align}
The images are represented using the bottom-up features proposed by \citet{anderson2018}. For each image, we extract a set of $R$ mean-pooled convolutional features $v_r \in \mathbb{R}^{I}$, one for each proposed image region $r$. We introduce $\bm{W}_3 \in \mathbb{R}^{J \times I}$, which projects the image features of a single region into the joint embedding space:
\begin{equation}
\label{eq:image_embeddings}
    \bm{v}^{e}_r = \bm{W}_3 \bm{v}_r
\end{equation}
To form a single representation $\bm{v}^*$ of the image from the set of embedded image region features $\bm{v}^{e}_r$, we apply a weighting mechanism. We generate a normalized weighting of region features $\beta  \in \mathbb{R}^{R}$ using $\bm{W}_4 \in \mathbb{R}^{1 \times J}$. $\beta_r$ denotes the weight for a specific region r. Then we sum the weighted region features to generate $\bm{v}^* \in \mathbb{R}^{J}$:
\begin{align}
    \beta_r' & = \bm{W}_4 \bm{v}^{e}_r \\
    \bm{\beta}  & = \text{softmax}(\bm{\beta}') \\
    \label{eq:image_representation}
    \bm{v}^* & = \sum_{r=1}^{R}{\beta_r
    \bm{v}^{e}_r}
\end{align}
We define the similarity between an image and a caption as the cosine similarity $cos(\bm{v}^*, \bm{s}^*)$. 

\subsection{Caption Generation}
For caption generation, we introduce a separate language generation LSTM that is stacked on top of the language encoding LSTM. At each timestep $t$, we first calculate a weighted representation of the input image features. We calculate a normalized attention weight $\alpha_t \in \mathbb{R}^{R}$ (one $\alpha_{r,t}$ for each region) using the language encoding and the image region features. Then, we create a single weighted image feature vector:
\begin{align}
    \alpha_{r,t}' & = \bm{W}_5 \text{tanh}(\bm{W}_6 \bm{v}_r^{e} + \bm{W}_7 h^l_t) \\
    \bm{\alpha}_{t} & = \text{softmax}(\bm{\alpha}_{r,t}') \\
    \hat{\bm{v}_t} & = \sum_{r=1}^{R}{\bm{\alpha}_{r,t} \bm{v}_r^{e}}
\end{align}
where $\bm{W}_5 \in \mathbb{R}^{H}$, $\bm{W}_6 \in \mathbb{R}^{H \times J}$ and $\bm{W}_7 \in \mathbb{R}^{H \times L}$. $H$ indicates the hidden layer dimension of the attention module.

These weighted image features $\hat{\bm{v}_t}$, the output of the language encoding LSTM $\bm{h}^l_{t}$  (Eqn. \ref{eq:lang_enc_lstm}) and the previous state of the language generation LSTM $\bm{h}^g_{t-1}$ are input to the language generation LSTM:
\begin{align}
    \label{eq:gen_lstm}
    \bm{h}^g_t = \text{LSTM}([\hat{\bm{v}_t}, \bm{h}^l_{t}], \bm{h}^g_{t-1})
\end{align}
The hidden layer dimension of the LSTM is $G$. The output probability distribution over the vocabulary is calculated using $\bm{W}_8 \in \mathbb{R}^{V \times G}$:
\begin{align}
    \label{eq:pred_word}
    p(w_t | w_{<t}) = \text{softmax}(\bm{W}_8 \bm{h}^g_t)
\end{align}


\subsection{Training}
The model is jointly trained on two objectives. The caption generation component is trained with a cross-entropy loss, given a target ground-truth sentence $s$ consisting of the words $w_1,\dots,w_{T}$:
\begin{equation}
    \mathcal{L}_{\text{gen}}(\theta_{1}) = - \sum_{t=1}^{T}{\text{log}\;p(w_t | w_{<t}; i)}
\end{equation}

The image--caption ranking component is trained using a hinge loss with emphasis on hard negatives \cite{faghri2017}:
\begin{multline}
    \mathcal{L}_{\text{rank}}(\theta_{2}) = \max_{s'}[\alpha+\text{cos}(i,s')-\text{cos}(i,s)]_+ \\ 
    + \max_{i'}[\alpha+\text{cos}(i',s)-\text{cos}(i,s)]_+
\end{multline}
where $[x]_+ \equiv max(x,0)$.

These two loss terms can take very different magnitudes during training, and thus can not be simply added. We use GradNorm \cite{chen2018} to learn loss weighting parameters $w_{gen}$ and $w_{rank}$ with an additional optimizer during training. These parameters dynamically rescale the gradients so that no task becomes too dominant. The overall training objective is formulated as the weighted sum of the single-task losses:
\begin{equation}
    \mathcal{L}(\theta_{1}, \theta_{2}) = w_{\text{gen}}\mathcal{L}_{\text{gen}}(\theta_{1}) + w_{\text{rank}}\mathcal{L}_{\text{rank}}(\theta_{2})
\end{equation}

\subsection{Inference}
The model generates $B$ captions for each image using beam search decoding. At each timestep, the tokens generated so far for each item on the beam are input back into the language encoder (Eqn. \ref{eq:projected_encoder_embeddings}). The output of the language encoder is concatenated with the image representation (Eqn. \ref{eq:image_representation}) and the previous hidden state of the generation LSTM, and input to the generation LSTM (Eqn. \ref{eq:gen_lstm}) to predict the next token (Eqn. \ref{eq:pred_word}).

The jointly-trained image--sentence ranking component can be used to re-rank the generated captions comparing  the image embedding with a language encoder embedding of the captions (Eqn. \ref{eq:image_embeddings}). We expect the ranking model will produce a better ranking of the $B$ captions than only beam search by considering their relevance and informativity with respect to the image.


\section{Results}

\begin{table}[t]
\centering
\begin{tabular}{@{}lccccc@{}} \toprule
\bf Model & \bf R & \bf M  & \bf S & \bf C & \bf B \\ \midrule
\textsc{sat} & 3.0 & 23.2 & 16.6 & 80.4 & 27.5 \\
\textsc{butd} & 6.5 & 25.8 & 19.1 & \textbf{98.1} & \textbf{32.6} \\
\textsc{\modelname} & 6.5 & 25.7 & 19.0 & 97.0 & 32.0  \\
\textsc{\modelname} + \textsc{rr} & \textbf{13.2} & \textbf{26.4} & \textbf{20.4} & 92.7 & 28.8 \\
\midrule
\textsc{full} & 33.3 & 27.4 & 20.9 & 105.3 & 36.6 \\
\bottomrule
\end{tabular}
\caption{\label{tab:results} Average results for Recall@5 (\textbf{R}; Eqn. \ref{eq:metric}), METEOR \cite[\textbf{M};][]{denkowski2014meteor}, SPICE \cite[\textbf{S};][]{anderson2016} , CIDEr \cite[\textbf{C};][]{Vedantam2015}, BLEU \cite[\textbf{B};][]{Papineni2002}. \textsc{rr} stands for re-ranking after decoding.}
\end{table}

We follow the experimental protocol defined in Section \ref{sec:experiments} to evaluate the joint model. See Appendix \ref{sec:training_details} for training details and hyperparameters.

Table \ref{tab:results} shows the compositional generalization performance, as well as the common image captioning metric scores for all models.
\modelname{} uses the same image features and a decoder architecture as the \textsc{butd} model. Thus, when using the standard beam search decoding method, \textsc{butr} does not improve over \textsc{butd}. However, when using the improved decoding mechanism with re-ranking \modelname{} + \textsc{rr}, Recall@5 increases to 13.2. We also observe an improvement in METEOR and SPICE, and a drop in BLEU and CIDEr compared to the other models. We note that BLEU has the weakest correlations \cite{elliott2014comparing}, and SPICE and METEOR have the strongest correlations with human judgments \cite{kilickaya2017re}.

\begin{table}[t]
\centering
\begin{tabularx}{0.48\textwidth}{lcccccc@{}}
\toprule
 & \multicolumn{2}{c}{Color} & \multicolumn{2}{c}{Size} & \multicolumn{2}{c}{Verb} \\
\cmidrule(rl){2-3} \cmidrule(rl){4-5} \cmidrule(rl){6-7}
& A & I & A & I & T & I \\
\midrule
\textsc{sat} & 3.7 & 10.5 & 0 & 0 & 1.6 & 2.2 \\
\textsc{butd} & 5.4 & 10.9 & 0.5 & 0 & 11.6 & 10.3 \\
\textsc{\modelname} & 6.4 & 16.2 & 0.3 & 0.2 & 7.0 & 8.6 \\
+ \textsc{rr} & \textbf{13.8} & \textbf{26.0} & \textbf{1.4} & \textbf{0.8} & \textbf{20.3} & \textbf{16.9} \\
\midrule
\textsc{full} &  42.7 & 38.7 & 5.9 & 33.3 & 39.6 & 39.5 \\
\bottomrule
\end{tabularx}
\caption{\label{tab:qualitative_results} Detailed Recall@5 scores for different categories of held out pairs. The scores are averaged over the set of scores for pairs from the respective category. \textsc{rr} stands for re-ranking after decoding. Color and size adjectives are split into \textbf{A}nimate or \textbf{I}nanimate objects; Verbs are split into \textbf{T}ransitive and \textbf{I}ntransitive verbs.}
\end{table}

\begin{figure*}[ht]
\center{\includegraphics[scale=0.42]
{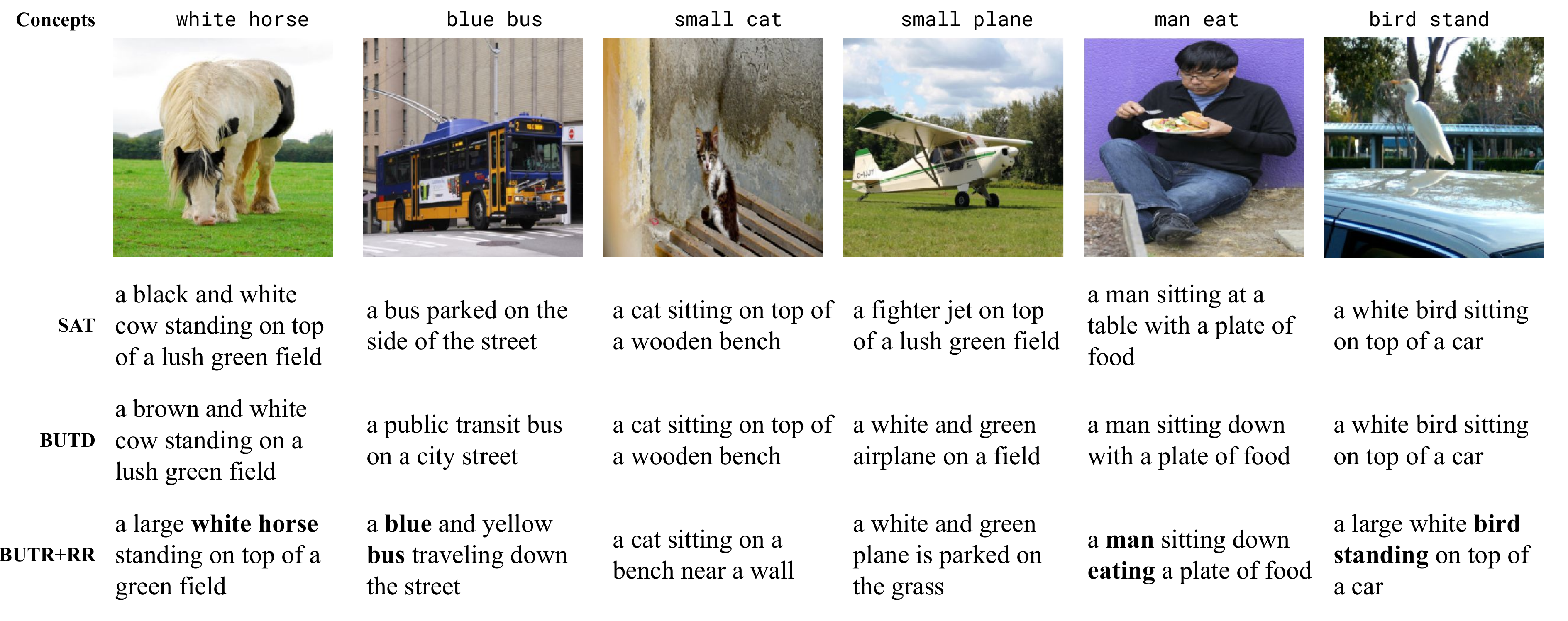}}
\caption{\label{fig:example_captions} Selected examples of the captions generated by \textsc{sat}, \textsc{butd}, and \modelname{} for six different concept pairs. The \textbf{bold words} in a caption indicate compositional success.}
\end{figure*}

The Recall@5 scores for different categories of held out pairs is presented in in Table \ref{tab:qualitative_results}, and Figure \ref{fig:example_captions} presents examples of images and the generated captions from different models. We observe that all models are generally best at describing colors, especially of inanimate objects; they nearly never correctly describe held out size modifiers; and for held out noun--verb pairs, performance is slightly better for transitive verbs.

\section{Analysis and Discussion}

\begin{table*}[ht]
\centering
\begin{tabular}{@{}lccccccc@{}}
\toprule
Model & ASL & Types & TTR$_1$ & TTR$_2$ & \%Novel & Cov & Loc$_5$ \\ \midrule
\citet{liu2017mat} & \textbf{10.3} $\pm$ 1.32 & 598 & 0.17 &  0.38 & 50.1 & 0.05 & 0.70 \\
\citet{vinyals2017show} & 10.1 $\pm$ 1.28 & 953 & 0.21 & 0.43 & 90.5 & 0.07 & 0.69 \\
\citet{shetty2017speaking} &  9.4 $\pm$ 1.31 & \textbf{2611} & 0.24 & 0.54 & 80.5 & \textbf{0.20} & 0.71 \\
\textsc{butd} & 9.0 $\pm$ 1.01 & 1162 & 0.22 & 0.49 & 56.4 & 0.09 & 0.78 \\
\modelname+\textsc{rr} & 10.2 $\pm$ \textbf{1.76} & 1882 & \textbf{0.26} & \textbf{0.59} & \textbf{93.6} & 0.14 & \textbf{0.80} \\ 
\midrule
Validation data & 11.3 $\pm$ 2.61 & 9200 & 0.32 & 0.72 & 95.3 & - & - \\
\bottomrule
\end{tabular}
\caption{\label{tab:diversity} Scores for diversity metrics as defined by \citet{van2018measuring} for different models.}
\end{table*}

\paragraph{Describing colors:}
The color--noun pairings studied in this work have the best generalization performance. We find that all models are better at generalizing to describing inanimate objects instead of animate objects, as shown in the detailed results in Table \ref{tab:qualitative_results}. One explanation for this could be that the colors of inanimate objects tend to have a higher variance in chromaticity when compared to the colors of animate objects \cite{rosenthal2018color}, making them easier to distinguish.

\paragraph{Describing sizes:} 
The generalization performance for size modifiers is consistently low for all models. The CNN image encoders are generally able to predict the sizes of object bounding boxes in an image. However, this does not necessarily relate to the actual sizes of the objects, given that this depends on their distance from the camera. To support this claim, we perform a correlation analysis in Appendix \ref{appendix:describing_sizes} showing that the bounding box sizes of objects in the COCO dataset do not relate to the described sizes in the respective captions.

Nevertheless, size modification is challenging from a linguistic perspective because it requires reference to an object's comparison class \cite{cresswell1977semantics, bierwisch1989semantics}. A large mouse is so with respect to the class of mice, not with respect to the broader class of animals. To successfully learn size modification, a model needs to represent such comparison classes.

We hypothesize that recall is reasonable in the \textsc{full} setting because it exploits biases in the dataset, e.g. that trucks are often described as \textsc{big}. In that case, the model is not actually learning the meaning of \textsc{big}, but simple co-occurrence statistics for adjectives with nouns in the dataset.

\paragraph{Describing actions:}
In these experiments, the models were better at generalizing to transitive verbs than intransitive verbs. This may be because images depicting transitive events (e.g. eating) often contain additional arguments (e.g. cake); thus they offer richer contextual cues than images with intransitive events. The analysis in Appendix \ref{appendix:describing_actions} provides some support for this hypothesis.

\paragraph{Diversity in Generated Captions:}\label{sec:diversity}
A crucial difference between human-written and model-generated captions is that the latter are less diverse \citep{Devlin2015language, dai2017towards}. Given that \modelname{}+\textsc{rr} improves compositional generalization, we explore whether the diversity of the captions is also improved.
\citet{van2018measuring} proposes a suite of metrics to measure the diversity of the captions generated by a model. We apply these metrics to the captions generated by \modelname{}+\textsc{rr} and \textsc{butd} and compare the scores to the best models evaluated in \citet{van2018measuring}.

The results are presented in Table \ref{tab:diversity}. \modelname{}+\textsc{rr} shows the best performance as measured by most of the diversity metrics. \modelname{}+\textsc{rr} produces the highest percentage of novel captions (\%Novel), which is important for compositional generalization. It generates sentences with a high average sentence length (ASL) -- performing similarly to \citet{liu2017mat} -- but with a larger standard deviation, suggesting a greater variety in the captions. The total number of word types (Types) and coverage (Cov) are higher for \citet{shetty2017speaking}, which is trained with a generative adversarial objective in order to generate more diverse captions. However, these types are more equally distributed in the captions generated by \modelname{}+\textsc{rr}, as shown by the higher mean segmented type-token ratio (TTR$_1$) and bigram type-token ratio (TTR$_2$).

The increased diversity of the captions may explain the lower BLEU score of \modelname{}+\textsc{rr} compared to \textsc{butd}. Recall that BLEU measures weighted n-gram precision, hence it awards less credit for captions that are lexically or syntactically different than the references. Thus, BLEU score may decrease if a model generates diverse captions. We note that METEOR, which incorporates non-lexical matching components in its scoring function, is higher for \modelname{}+\textsc{rr} than \textsc{butd}.

\paragraph{Decoding strategies:} The failure of the captioning models to generalize can be partially ascribed to the effects of maximum likelihood decoding. \newcite{holtzman2019curious} find that maximum likelihood decoding leads to unnaturally flat and high per-token probability text. We find that even with grounding from the images, the captioning models do not assign a high probability to the sequences containing compositions that were not observed during training.
\modelname{} is jointly trained with a ranking component, which is used to re-rank the generated captions, thereby ensuring that at the sentence-level, the captions are relevant for the image. It can thus be viewed as an improved decoding strategy such as those proposed in \newcite{vijayakumar2018diverse, fan2018hierarchical,radford2019language, holtzman2019curious}.

\section{Conclusion}

Image captioning models are usually evaluated without explicitly considering their ability to generalize to unseen concepts. In this paper, we argued that models should be capable of \textit{compositional generalization}, i.e. the ability to produce captions that include combinations of unseen concepts. We evaluated the ability of models to generalize to unseen adjective--noun and noun--verb pairs and found that two state-of-the-art models did not generalize in this evaluation, but that an image--sentence ranking model did. Given these findings, we presented a multi-task model that combines captioning and image--sentence ranking, and uses the ranking component to re-rank the captions generated by the captioning component. This model substantially improved generalization performance \textit{without} sacrificing performance on established text-similarity metrics, while generating more diverse captions. We hope that this work will encourage researchers to design models that better reflect human-like language production.

Future work includes extending the evaluation to other concept pairs and other concept classes, analysing the circumstances in which the re-ranking step improves compositional generalization, exploring the utility of jointly trained discriminative re-rankers into other NLP tasks, developing models that generalize to size modifier adjectives, and devising approaches to improve the handling of semantically equivalent outputs for the proposed evaluation metric.

\section*{Acknowledgements}

We thank Emiel van Miltenburg and Ákos Kádár for their extensive feedback on the work, and the reviewers, Ana Valeria Gonzales, Daniel Hershcovich, Heather Lent, and Mareike Hartmann for their comments. We also thank the participants of the Lorentz Center workshop on Compositionality in Brains and Machines for suggesting the phrase ``paradigmatic gap''. MN was supported by the Erasmus+ Traineeship program. RA and MA are funded by a Google Focused Research Award.


\bibliographystyle{acl_natbib}

\begin{appendix}
\clearpage
\appendix

\section{Adjective Frequencies}\label{appendix:adjective_frequencies}

\begin{figure}[!htb]
\center{\includegraphics[width=0.45\textwidth]
{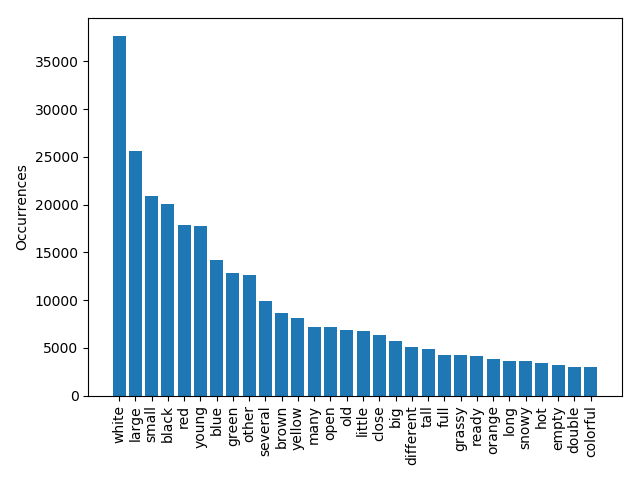}}
\caption{\label{fig:adjectives} Histogram of the adjectives used in COCO.}
\end{figure}

Figure \ref{fig:adjectives} shows a histograme of the most frequent adjectives in the captions of the COCO dataset.

\section{Concept Pairs Statistics}\label{appendix:held_out_pairs}

Table \ref{tab:concept_pairs_statistics} shows the number of images for which at least one reference caption includes the respective concept pair. The two numbers indicate scores for the COCO training set (which is also used for training, by holding out exactly this set of images) and the COCO validation set (which is used for evaluation).

\begin{table}[!ht]
\centering
\begin{tabular}{lrr}
\toprule
& Training Set & Validation Set \\
\midrule
\small{\texttt{black bird}} & 205 & 122  \\
\small{\texttt{small dog}} & 681 & 316  \\
\small{\texttt{white boat}} & 373 & 196 \\ 
\small{\texttt{big truck}} & 417 & 191  \\
\small{\texttt{eat horse}} & 212 & 106 \\
\small{\texttt{stand child}} & 1288 & 577 \\ 
\small{\texttt{white horse}} & 264 & 151  \\
\small{\texttt{big cat}} & 184 & 103  \\
\small{\texttt{blue bus}} & 276 & 143  \\
\small{\texttt{small table}} & 261 & 134  \\
\small{\texttt{hold child}} & 1328 & 664  \\
\small{\texttt{stand bird}} & 532 & 260 \\
\small{\texttt{brown dog}} & 613 & 291 \\
\small{\texttt{small cat}} & 252 & 149 \\
\small{\texttt{white truck}} & 262 & 121  \\
\small{\texttt{big plane}} & 967 & 357  \\
\small{\texttt{ride woman}} & 595 & 300  \\
\small{\texttt{fly bird}} & 245 & 132  \\
\small{\texttt{black cat}} & 840 & 448  \\
\small{\texttt{big bird}} & 215 & 123 \\
\small{\texttt{red bus}} & 566 & 232  \\
\small{\texttt{small plane}} & 481 & 158  \\
\small{\texttt{eat man}} & 555 & 250  \\
\small{\texttt{lie woman}} & 301 & 144  \\
\bottomrule
\end{tabular}
\caption{\label{tab:concept_pairs_statistics} Number of occurrences of concept pairs in the COCO training and validation set. The full training set size is 82,783 images, the validation set consists of 40,504 images.}
\end{table}
\section{Dataset splits}\label{appendix:dataset_splits}
To increase the efficiency of training and evaluation, we create training sets in which we simultaneously hold out multiple pairs. We ensure that no more than 5\% of the training data is removed from the original training set, and that we do not remove pairs with overlapping nouns, adjectives or verbs within the same training set.

Based on these constraints, we create four sets of training and evaluation splits. Each set contains a held out pair for a color modifier on an animate and inanimate object, a size modifier on an inanimate and inanimate object and a transitive and an intransitive verb for animate objects. For each of these four splits, we train a model on the respective training data and calculate the recall for each held out pair on the respective evaluation data.

Further, we calculate average recall scores for various groups of conceptually similar held out pairs and an average over all recall scores as a single measure indicating the compositional generalization performance of a model.

Table \ref{tab:dataset_splits} lists the held out word pairs and their distribution into four different datasets. We did not include inanimate verb--noun pairs because there were not enough instances in the validation set.

\begin{table}[ht!]
\centering
\begin{tabularx}{.5\textwidth}{@{}p{0.05\linewidth}Xp{0.15\linewidth}p{0.15\linewidth}@{}}
\toprule
& \bf Held out pairs & \bf $\mathcal{D}_\text{train}$ & \bf $\mathcal{D}_\text{eval}$\\
\midrule
1 & \small{\texttt{black cat}, \texttt{big bird}, \texttt{red bus}, \texttt{small plane}, \texttt{eat man}, \texttt{lie woman}} & 79,825 & 1,355 \\ 
2 & \small{\texttt{brown dog}, \texttt{small cat}, \texttt{white truck}, \texttt{big plane}, \texttt{ride woman}, \texttt{fly bird}}  & 79,849 & 1,350 \\
3 & \small{\texttt{white horse}, \texttt{big cat}, \texttt{blue bus}, \texttt{small table}, \texttt{hold child}, \texttt{stand bird}} & 79,938 & 1,455 \\
4 & \small{\texttt{black bird}, \texttt{small dog}, \texttt{white boat}, \texttt{big truck}, \texttt{eat horse}, \texttt{stand child}} & 79,607 & 1,508\\
\bottomrule
\end{tabularx}
\caption{\label{tab:dataset_splits} The held out word pairs in each dataset split. Training and evaluation set sizes are in number of images; each image is associated with five captions. The full training set size is 82,783 images.}
\end{table}

\section{Synonyms}\label{appendix:synonyms}

Table \ref{tab:synonyms} shows the synonyms we defined for our selected adjectives and verbs. For the noun synonyms, refer to \citet[Appendix]{lu2018neural}

\begin{table}[ht!]
\centering
\begin{tabularx}{.5\textwidth}{@{}p{0.14\linewidth}X@{}}
\toprule
\textbf{Word} & \textbf{Synonyms} \\
\midrule
big & large, tall, huge, wide, great, broad, enormous, expansive, extensive, giant, gigantic, massive, vast \\
small & little, narrow, short, tinier, tiny, thin, compact, mini, petite, skinny\\
red & dark-red, light-red\\
brown & brownish, dark-brown, light-brown\\
blue & blueish, light-blue, dark-blue \\
black & - \\
white & - \\
\midrule
eat & chew, bite, graze \\
lie & lay \\
hold & carry\\
ride & - \\
fly & - \\
stand & - \\
\bottomrule
\end{tabularx}
\caption{\label{tab:synonyms} The adjective and verb synonyms used to select word pairs for the experiments in this paper.}
\end{table}

\section{Training \modelname{}}\label{sec:training_details}
In this section we describe the hyperparameters and training details of \modelname{}. The parameters have been chosen in accordance with the \textsc{butd} and VSE++ models and not further tuned. \modelname{} is trained with a 1024D visual-semantic embedding space ($J$), a 1000D language encoding LSTM ($L$), a 1000D language generation LSTM ($G$), a vocabulary of 10000 types ($V$), 300D word embeddings ($E$), 2048D image region feature vectors, a 512D attention model dimension, and inference is performed using beam search with a 100 hypotheses ($B$). \modelname{} is trained using pre-computed bottom-up image features from 36 regions obtained using the bottom-up encoder defined in \citet{anderson2016}. The caption generation component is trained with teacher forcing and a maximum caption length of 20 in batches of 100 with the Adam optimizer \cite{kingma2014adam} using an initial learning rate of 1e-4. The gradients are clipped when they exceed 10.0. For the GradNorm optimizer, we also use Adam, but with an initial learning rate of 0.01. We set the asymmetry to 2.5. \textsc{butr} is trained for at most 30 epochs, and early stop when the validation set BLEU score does not increase for five consecutive epochs.

\section{Describing Sizes}\label{appendix:describing_sizes}
To support the claim that the bounding box sizes do not necessarily relate to the actual sizes of the objects as they are described, we perform a correlation analysis. We make use of the fact that there is bounding box annotations for objects in the COCO dataset. We identify each noun concept that was also used in combination with size modifiers in the held out concept pairs (cf. Table \ref{tab:task:word_pairs}: \texttt{cat}, \texttt{plane}, \texttt{table}, \texttt{dog}, \texttt{bird}, and \texttt{truck}. For each of these concepts, we consider all images that contain at least one instance of the object as annotated in the COCO dataset. Given one of these images, we regard only the size of the area of the biggest bounding box\footnote{We assume that the biggest object of a category in the image is also the most salient and thus most likely the one that was described.} belonging to an object of that kind. Then, we look at the reference captions belonging to the respective image and look for matching concept pairs\footnote{We disregard all images with contradicting descriptions (i.e. different annotators describe the object as \texttt{small} and \texttt{big}) and images where the size of the concept is not described at all.}. To test whether the bounding box sizes relate to the described sizes of the objects, we perform a unpaired t-test comparing the box sizes for objects described as \texttt{small} and objects described as \texttt{big}.

\begin{table*}[ht]
\centering
\begin{tabular}{@{}lcccc@{}}
\toprule
Concept & Average bounding box size (in pixels)& Number of samples & p-Value \\ \midrule
\texttt{small cat} & 42,920.6 $\pm$ 38,952.2 & 628 & \multirow{2}{*}{0.64} \\
\texttt{big cat} & 44,057.4 $\pm$ 41,979.9 & 495 \\
\midrule
\texttt{small plane} & 33,718.8 $\pm$ 30,481.2 & 569 & \multirow{2}{*}{0.77} \\
\texttt{big plane} & 33,263.1 $\pm$ 31,722.9 & 1,408 \\
\midrule
\texttt{small dog} & 36,939.5 $\pm$ 41,073.3 & 1,109 & \multirow{2}{*}{0.94} \\
\texttt{big dog} & 37,098.3 $\pm$ 40,088.6 & 718 \\
\midrule
\texttt{small table} & 80,762.0 $\pm$ 89,751.0 & 1,860 & \multirow{2}{*}{0.007} \\
\texttt{big table} & 72,958.0 $\pm$ 91,340.0 & 2,101 \\
\midrule
\texttt{small bird} & 15,063.0 $\pm$ 19,487.6 & 774 & \multirow{2}{*}{0.77} \\
\texttt{big bird} & 14,707.8 $\pm$ 27,008.7 & 789 \\
\midrule
\texttt{small truck} & 30,014.0 $\pm$ 49,121.4 & 531 & \multirow{2}{*}{0.21} \\
\texttt{big truck} & 32,918.2 $\pm$ 46,379.8 & 1,945\\
\bottomrule
\end{tabular}
\caption[Comparison of bounding box sizes for different concept pairs describing sizes of objects.]{\label{tab:object_sizes} Comparison of bounding box sizes for different concept pairs describing sizes of objects. The last column indicates the resulting p-value from an unpaired t-test between the data of the two respective rows.}
\end{table*}

Table \ref{tab:object_sizes} shows the average bounding box size for the set of concept pairs. Further, the last column shows the resulting p-values from the t-tests. The differences in box sizes for \texttt{small} vs. \texttt{big} objects are never significant, except for the case of \texttt{table} (p $\approx$ 0.007). However, in this case the box sizes are on average bigger if the \texttt{table} is described as \texttt{small}. We conclude that the bounding box sizes of objects in the COCO dataset do not relate to the described sizes in the respective captions.

\section{Describing Actions}\label{appendix:describing_actions}

\begin{table}[htb]
\centering
\begin{tabularx}{.48\textwidth}{@{}lcc}
\toprule
Concept Pair & with Object & including "obl" \\ \midrule
\texttt{hold child} & 96\% & 99\% \\
\texttt{ride woman} & 81\% & 97\% \\
\texttt{eat man} & 87\% & 97\% \\
\texttt{stand child} & 26\% & 92\% \\
\texttt{stand bird} & 3\% & 98\% \\
\texttt{fly bird} & 7\% & 89\% \\
\texttt{lie woman} & 24\% & 96\% \\
\bottomrule
\end{tabularx}
\caption[Percentage of captions where a direct or indirect object is connected to the noun of the concept pair.]{\label{tab:pairs_objects} Percentage of captions where a direct or indirect object is connected to the noun of the concept pair. In the last column, additional arguments ("obl") are also counted as objects.}
\end{table}

We analyze the dataset and calculate statistics on the occurrence of objects in connection with the concept pairs that include transitive and intransitive verbs. We use StanfordNLP for detecting the objects. The examined concept pairs for transitive verbs are \texttt{hold child}, \texttt{ride woman}, \texttt{eat man} and for intransitive verbs \texttt{stand child}, \texttt{stand bird}, \texttt{fly bird}, and \texttt{lie woman}.\footnote{We exclude the pair \texttt{eat horse} from the analysis, because we defined "graze" as a synonym for "eat" (cf. Table \ref{tab:synonyms} which is an intransitive verb. We find that this is quite often used and thus would decrease the validity of the statistics}

The results are presented in Table \ref{tab:pairs_objects}. In fact, phrases using transitive verbs contain objects 88\% of the time and phrases using intransitive verbs only 15\% of the time. If we include additional arguments (marked as oblique "obl") in our definition of objects, the percentage in the transitive verb case rises to 98\%, and in the intransitive case to 93\%. An unpaired t-test shows that this difference is still significant (p~$< 10^{-38}$).

The performed analysis supports the hypothesis that the models perform better for actions described with transitive verbs because of additional clues coming from the object.

\section{Detailed Results}

Table \ref{tab:extensive_results} presents the Recall@5 generalization performance for each held out pair.

\begin{table*}[t]
\centering
\begin{tabularx}{.68\textwidth}{p{0.135\linewidth}rrrrr}
\toprule
& \textbf{\textsc{sat}} & \textbf{\textsc{butd}} & \textbf{\modelname{}} & \textbf{\textsc{+rr}} & \textbf{\textsc{full}} \\
\midrule
\small{\texttt{black bird}} & 7.4 & 1.6 & 4.1 & \textbf{9.8} & 25.4  \\
\small{\texttt{small dog}} & 0 & \textbf{0.3} & 0 & \textbf{0.3} & 13.0  \\
\small{\texttt{white boat}} & 1.5 & 5.1 & 4.6 & \textbf{8.2} & 17.3 \\ 
\small{\texttt{big truck}} & 0 & 0 & 0 & \textbf{0.5} & 35.1  \\
\small{\texttt{eat horse}} & 0 & 19.8 & 7.5 & \textbf{36.8} & 41.5  \\
\small{\texttt{stand child}} & 0.7 & 3.6 & 3.1 & \textbf{14.0} & 24.4 \\ 
\small{\texttt{white horse}} & 4.0 & 10.6 & 9.9 & \textbf{13.9} & 48.3  \\
\small{\texttt{big cat}} & 0 & 0 & 0 & 0 & 0  \\
\small{\texttt{blue bus}} & 15.4 & 6.3 & 22.4 & \textbf{28.0} & 40.6  \\
\small{\texttt{small table}} & 0 & 0 & 0 & 0 & 0.7  \\
\small{\texttt{hold child}} & 3.2 & 5.9 & 3.2 & \textbf{11.6} & 33.7  \\
\small{\texttt{stand bird}} & 1.2 & 6.9 & 5.8 & \textbf{11.2} & 41.2 \\
\small{\texttt{brown dog}} & 0.3 & 1.4 & 3.8 & \textbf{9.3} & 29.9 \\
\small{\texttt{small cat}} & 0 & 0 & \textbf{1.3} & \textbf{1.3} & 0.7 \\
\small{\texttt{white truck}} & 8.3 & 8.3 & 8.3 & \textbf{19.0} & 31.4  \\
\small{\texttt{big plane}} & 0 & 0 & 0.8 & \textbf{2.5} & 58.3  \\
\small{\texttt{ride woman}} & 0 & 10.7 & 3.7 & \textbf{15.3} & 46.0  \\
\small{\texttt{fly bird}} & 6.1 & 19.7 & 21.2 & \textbf{25.0} & 52.3  \\
\small{\texttt{black cat}} & 3.1 & 7.8 & 7.8 & \textbf{22.3} & 67.2  \\
\small{\texttt{big bird}} & 0 & 1.6 & 0 & \textbf{4.1} & 9.8 \\
\small{\texttt{red bus}} & 16.8 & 24.1 & 29.7 & \textbf{48.7} & 65.5  \\
\small{\texttt{small plane}} & 0 & 0 & 0 & 0 & 39.2  \\
\small{\texttt{eat man}} & 3.2 & 10 & 13.6 & \textbf{17.6} & 37.2  \\
\small{\texttt{lie woman}} & 0.7 & 11.1 & 4.2 & \textbf{17.4} & 40.3  \\
\bottomrule
\end{tabularx}
\caption[Recall@5 for each of the held out concept pairs.]{\label{tab:extensive_results} Recall@5 for each of the held out concept pairs. \textsc{rr} stands for re-ranking after decoding. The \textbf{bold face} results denote the best model performance when trained with paradigmatic gaps.}
\end{table*}

\end{appendix}

\end{document}